%% file: main.tex
\documentclass[10pt]{article} 
\usepackage[preprint]{tmlr}

\input{additional-pkg}

\title{Agentics 2.0: Logical Transduction Algebra for Agentic Data Workflows}

\author{\name Alfio Massimiliano Gliozzo \thanks{Equal contribution.} \email gliozzo@us.ibm.com \\
      \addr IBM
      \AND
      \name Junkyu Lee \footnotemark[\value{footnote}] \email junkyu.lee@ibm.com \\
      \addr IBM
      \AND
      \name Nahuel Defosse \email nahuel.defosse@ibm.com\\
      \addr IBM}

\begin{document}

\maketitle

\begin{abstract}
Agentic AI is rapidly transitioning from research prototypes to enterprise deployments, 
where requirements extend to meet the software quality attributes of reliability, scalability, and observability beyond plausible text generation. 
We present \texttt{Agentics 2.0}, a lightweight, Python-native framework for building high-quality, structured, explainable, and type-safe agentic data workflows.
At the core of \texttt{Agentics 2.0}, the logical transduction algebra formalizes a large language model inference call as a typed semantic transformation, which we call a transducible function
that enforces schema validity and the locality of evidence.
The transducible functions compose into larger programs via algebraically grounded operators and execute as stateless asynchronous calls in parallel in asynchronous Map-Reduce programs.
The proposed framework provides semantic reliability through strong typing, semantic observability through evidence tracing between slots of the input and output types, and scalability through stateless parallel execution.
We instantiate reusable design patterns and evaluate the programs in Agentics 2.0 on challenging benchmarks, including DiscoveryBench for data-driven discovery and Archer for NL-to-SQL semantic parsing, demonstrating state-of-the-art performance.
\end{abstract}

\input{introduction}
\input{lta}
\input{agentics20}
\input{discoverybench}
\input{archer}
\input{conclusion}

\bibliography{ref}
\bibliographystyle{tmlr}

\appendix
\input{supp}

\end{document}

%% file: additional-pkg.tex
\usepackage{amsmath}
\usepackage{amsthm}
\newtheorem{example}{Example}

\usepackage{times}
\usepackage{soul}
\usepackage[utf8]{inputenc}
\usepackage{hyperref}
\usepackage{url}

\usepackage[small]{caption}
\usepackage{subcaption} 
\usepackage{graphicx}
\usepackage{booktabs}
\usepackage{algorithm}
\usepackage{algorithmic}

\usepackage[switch]{lineno}
\usepackage{listings}
\usepackage{xcolor} 

\usepackage{ragged2e}
\usepackage{listings}
\DeclareCaptionStyle{ruled}{labelfont=normalfont,labelsep=colon,strut=off,justification=raggedright}
\floatstyle{plain}
\newfloat{listing}{tb}{lst}
\floatname{listing}{Listing}
\usepackage{booktabs}
\usepackage{siunitx}
\usepackage{tabularx} 

%% file: introduction.tex
\section{Introduction}
Agentic AI systems are rapidly transitioning from prototypes to production across real-world enterprise domains \citep{wang2024survey,guo2024large}, yet most current agentic data workflows still revolve around prompt chaining, state-graph orchestration, or schema-enforced function calls \citep{dao2025agentic}. These approaches struggle to deliver reliability, observability, and scalability  at the levels required by enterprise workloads
\footnote{The standard quality metrics can be found in the standardized vocabulary \citep{8016712}.}. 
Recent study confirms that practical systems significantly constrain autonomy and rely heavily on persistent human oversight \citep{pan2025measuring}.

We argue that these limitations stem from an agent-centric, anthropomorphic perspective that views the large language model (LLM) inference call as a conversation over agents with personas and roles. We can see two common patterns: the first is conversational agents that rely on black-box planners to generate and execute multi-step workflows through natural-language prompt chains, lacking verifiable control flows. The second is mutli-agent system that divides the task into sub-tasks in natural language and executes and coordinates via unreliable conversations.
Prior work on agentic workflow has shown that casting the LLM inference call into typed function calls improves the reliability and reusability of the agentic AI programs \citep{gliozzo2025transduction,pydanticai,mellea,khattab2024dspy}. 
However, without function-level semantics,  programs are limited in composing reliable workflows.

In this paper, we propose a principled programming model for LLM transduction inference that (1) constrains input and output with frames or semantic types, (2) composes transductions as functions in programming languages, and (3) preserves the semantic evidence that explains how the output semantic types are transduced from the input type.  
We formalize LLM transduction as a typed, composable, and evidence-preserving function algebra, which we call it as a transducible function. From the proposed formalism, the reliability becomes a software quality attribute because every step of LLM transduction must be a well-typed contract between input and output types. Namely, any ill-formed outputs trigger an error rather than a silent corruption of the text. Evidence requirements in transduction that explain how the slots in the input type are mapped to the output type prevent hallucinated slot fillings. 
Software observability is typically tracked at the level of LLM inference calls. 
In the proposed framework, it can be elevated to the level of semantics. 
Scalability can also be achieved in a high-level program by writing the agentic data workflow as a composition of stateless transductions, instead of deligating it to lower-level orchestrations of LLM API invocations.

This paper contributes:
(1) a typed, composable function algebra for LLM computation by formalizing transducible functions,
(2) agentic AI programming model that interleaves deterministic code and LLM transductions via typed asynchronous coroutines,
(3) semantic observability through the provenance and confidence across composed workflows, and
(4) empirical validation across challenging agentic AI benchmarks.

%% file: lta.tex
\section{Logical Transduction Algebra}
\subsection{Background}
Existing agentic AI programming frameworks offer workflows modeled with types and structured schemas. \citeauthor{gliozzo2025transduction} proposed a logical transduction algebra that formulates transduction as schema-constrained LLM inference. 
Conceptually, logical transduction is an abstraction of LLM-based transformation of frames or semantic types. 
This approach maps structured input objects to another output typed object,
and it allows asynchronous execution across a list of objects of the same type. 
Shifting the focus from conversation to transduction has improved the reliability of the pipeline by validating the types of LLM-generated text. 
Additionally, it supports a Map-Reduce style programming model for scaling out batch tasks.

Yet, the schema-constrained transduction still presents a conceptual gap in a multi-step pipeline that composes state-to-state transductions, since it does not account for the type contracts and the evidence of generations across transductions.
The absence of function composition semantics affects completeness and diminishes software quality, making it difficult to manage the accumulation of semantic errors that pass syntactic tests.

\subsection{Types as Semantically Grounded Objects}
A type in the logical transduction algebra is a finite record of named slots, each associated with specific types, realized in practice as Pydantic models \citep{gliozzo2025transduction}. 
The semantics of a type arise not only from syntactic constraints but also from natural language descriptions attached to its fields, which guide LLMs to be aware of the meanings of the type during structured decoding. 
Thus, types serve as executable specifications for input-output contracts. 
They validate generated instances, ground parameterized prompts, and act as a carrier of meaning throughout the workflow.

Formally, A type $T$ is defined recursively as 
$T:= b \mid [ s_1: T_1, \ldots, s_n: T_n ],$
where $b$ can be a basic type in programming languages 
$\{\texttt{str}, \texttt{int}, \texttt{float}, \texttt{bool}, \dots\}$.
We use lowercase letters to denote an instance or state of a type, e.g., $t$ is a state of $T$.
We denote the $i$-th slot of $T$ by $T_i$, 
and 
a subset of slots of $T$ 
by $T_S$ with $S \subset \{s_1, s_2, \ldots, s_n\}$.

Basic type operators enable the construction of composite states using the following operations on types.
\begin{itemize}
    \item \textbf{Merge}: Combines slots of two types $X$ and $Y$ as 
    $X \& Y$ that contains all the slots from both. 
    In the presence of conflicts when merging two instances $x$ and $y$, we resolve them by preferring one side.
    \item \textbf{Projection}: Projects type $X$ onto a new type $Y$ 
    by restricting the slots in $X$ to a subset $T_S$ as $Y = \Downarrow_{T_S}X$.
    We abuse the notation $X_S = \Downarrow_{T_S}X$ when it is clear from context.
    \item \textbf{Composition}: Forms a tuple of the source $X$ and target $Y$ types,
    $X \otimes Y = \big(X, Y\big)$.
\end{itemize}
These operations make intermediate representations explicit and align with function interfaces in the next subsection.

\subsection{Transducible Functions}
In this paper, we introduce a transducible function as a typed semantic transformation 
$f: X \rightarrow Y$ that satisfies the following four properties: 
\begin{itemize}
    \item \textbf{Typed}: $f$ returns a well-typed state $y \in Y$.
    \item \textbf{Explainability}: 
    Given states $x$ and $y$, there exists an explanation $e = \varphi(x, y)$.
    \item \textbf{Local Evidence}: A non-empty subset of input slots $X_S$ was used to compute the value $y_i$.
    \item \textbf{Provenance}: For all slots $Y_i \in Y$, there exists a mapping that records the slot $Y_i$ to local evidence $X_S$.
\end{itemize}
Transducible functions extend computable functions 
by requesting explicit explainability, the locality of evidence, and provenance maps that connect each output slot to specific input slots.
Note that any deterministic program satisfying these properties is also a transducible function. In addition, transducible functions can be realized by LLMs provided they satisfy the above properties.

\begin{example}
\normalfont
Let's take an example from credit risk assessment. 
The input type $X$ models a loan application, and the output type $Y$ is a credit evaluation.
Given an input state $x$, 
consider a transducible function $f$ returns $y=f(x)$ 
with the explanation and provenance map for $Y$.
Notice that the slot \texttt{last\_name} has not been taken into consideration for the explanation. 

\begin{lstlisting}[caption={Credit Risk Assessment Example}]
# Type declarations
X: [last_name: str, income: float, debt: float, credit_history: str],
Y: [risk_score: int,risk_class: [low, medium, high]],
# Instance of X and Y
x: {last_name: Smith, income: 60000, debt: 25000, credit_history: 'late payment in 2021'}
y = f(x)
  - y: {risk_score: 62, risk_class: medium}
  - explanation: 'The debt-to-income ratio is moderate, and a historical late payment increases risk'
  - provenance: [income, debt, credit_history]
\end{lstlisting}
\end{example}

\citeauthor{gliozzo2025transduction} introduces a logical transduction operator $\ll$ as a LLM-based transformation from a typed object $x$ to $y$, $Y \ll X$. 
In this paper, we propose to extend the operator $\ll$ to denote the class of all transducible functions mapping from type $X$ to $Y$.
Given concrete types $X$ and $Y$ and a natural language instruction string $I$, 
we can view $\ll$ as 
a realization of transducible functions selected by an LLM  that is the most consistent with $I$.
From the perspective of logical transduction algebra, the internal behavior of LLMs can remain as a black box as long as the external type contract and properties of the transducible function are respected. 

\subsection{Algebraic Structure of Transducible Functions}
Transducible functions form an algebra under standard operations whose objects are types and whose morphisms are transductions.

\paragraph{Identity and Associativity.}
For every type $X$, there exists an identity transduction
$$
I_X : X \ll X, \quad I_X(x) = x,
$$
which uses all slots of $X$ as evidence and satisfies 
$f \circ I_X = f = I_Y \circ f$ for all $f : Y \ll X$.

\paragraph{Composition.}
Given $f_1 : Y \ll X$ and $f_2 : Z \ll Y$, their sequential composition 
can be defined as,
\[
f_2 \circ f_1 : Z \ll X, \quad (f_2 \circ f_1)(x) = f_2(f_1(x)).
\]
We can show that transducible functions are closure under composition.
Namely, the composed mapping is itself transducible. 
Its evidence subset is obtained by chaining the evidence of $f_1$ and $f_2$,
and provenance traces can be composed accordingly.

Composition is associative,
$
f_3 \circ (f_2 \circ f_1) = (f_3 \circ f_2) \circ f_1,
$
so the set of all transducible functions with composition and identities forms a monoid.

\paragraph{Canonical Type Transductions.}
We extend the concept of transduction to the types,
namely, types themselves induce transductions. 
Let $G$ be a generic input type such as unstructured or weakly structured data, e.g., text, JSON fragments, tables, or multimodal signals. 
For each type $T$, there is a canonical constructor
$
\varphi_T : T \ll G,
$
which acts as a schema-constrained semantic parser from the generic input $G$ to the structured type $T$. 
Then, typing becomes a form of
computation. Given $x \in G$, the expression $T(x) = (T \ll G)(x)$ parses and validates $x$ into a
well-typed instance of $T$. 

\subsection{Map-Reduce Semantics}
The transducible functions can be applied in parallel using the Map-Reduce programming model.
Let $X^N$ denote a sequence of $N$ states of type $X$, $[x_1, \dots, x_N]$.

\paragraph{Map.}
Given a transducible function $f: Y \ll X$ and a collection 
$X^N = [x_1, \dots, x_N]$ with $x_i \in X$, 
$map$ applies $f$ pointwise,
\[
  \mathrm{map}(f) : Y^N \ll X^N,
\]
\[
  \mathrm{map}(f)([x_1, \dots, x_N]) = [f(x_1), \dots, f(x_N)].
\]
Each element is processed independently, 
preserving per-element evidence and provenance maps. 
$map$ can be processed in parallel and is well-suited to asynchronous execution.

\paragraph{Reduce.}
Let $r : Z \ll Y^N$ be a transducible function that aggregates a collection $Y^N$ 
into a single output $z \in Z$, 
e.g., voting, averaging, consensus formation, or global summarization. 
Because $r$ is itself transducible, its output carries an evidence subset and a
provenance mapping that links each part of $z$ to the contributing elements of 
the collection $Y^N$,
\[
  r : Y \ll X^N,
\]
\[
  r([x_1, \dots, x_N]) = z.
\]

\paragraph{Map-Reduce.}
A full $map$-$reduce$ computation can be obtained by composition,
\[
  r \circ \mathrm{map}(f) : 
  r \circ \big(Y^N \ll X^N\big) = 
  (Z \ll Y^N) \circ \big(Y^N \ll X^N\big)
  =  Z \ll X^N,
\]
which is again a transducible function from a collection of inputs $X^N$ to a single
aggregate output in $Z$. 
The algebra is also closed under these $map$ and $reduce$ operators. 
Note that asynchronous independence in the $map$ phases and structured aggregation in the $reduce$ phases ensure scalability.
The locality of evidence and the provenance property guarantee that explainability is preserved across the entire composed pipeline.

%% file: agentics20.tex
\section{Agentics 2.0}
\label{sec:patterns}

Agentics 2.0\footnote{
The framework provides native support for integrating LLMs, MCP tools, and structured data sources such as databases, JSON, and CSV files. It also offers built-in vector search and clustering functionality.}
is a Python software library that implements the logical transduction algebra, extended with transducible functions. 
The core design principle is to extend standard Python abstractions
such as Pydantic types and asynchronous functions with new operators that make logical transduction a first-class citizen in the language.

Concretely, Agentics 2.0 overloads operators such as $\ll$ (transduction),
\texttt{@} (type composition), and \texttt{\&} (type merge) so that Pydantic types, 
instances, and asynchronous functions can be combined into larger transducible workflows. 
In the rest of this section, we first introduce these language primitives and illustrate basic syntax and examples.
Once imported, Agentics 2.0 extends the behavior of Pydantic models and asynchronous functions by
introducing a small set of primitives.

\subsection{The \texorpdfstring{$\ll$}{<<} Operator}

The operator $\ll$ generates transducible functions from a pair of types. 
In Agentics 2.0, any pair of Pydantic types \texttt{X} and \texttt{Y} can be combined into a transducible function $Y \ll X$, which is stateless and async callable.
\begin{lstlisting}[language=Python]
from pydantic import BaseModel

class Question(BaseModel):
    prompt: str
    options: list[str]

class Answer(BaseModel):
    options: list[str]
    choice: str

# Create a transducible function "decide" 
decide = Answer << Question
\end{lstlisting}

Once a transducible function \texttt{decide} is created, the transduction can be invoked like any other async function.
\begin{lstlisting}[language=Python]
q = Question(
    prompt="When an electric motor has bearing damage , which sensor should be monitored?",
    options=["axial flux", "vibration", "voltage", "cooling gas", "partial discharge"])

answer = await decide(q)
print(answer.choice)
\end{lstlisting}

Internally, \texttt{decide} triggers an LLM-backed stateless agent that executes the transduction and returns a well-typed \texttt{Answer} state. 
The transduction parameters, such as instructions, tools,
model and temperature can be configured by wrapping them in an instance of type \texttt{With}.
\begin{lstlisting}[language=Python]
decide = (Answer << With(
    Question,
    instructions=(
        "Pick the best option and justify it using evidence from the prompt and your domain knowledge."
    ),
    tools=[sensor_knowledge_base],     
    model="gpt-oss-120"), 
    explanation=True)
\end{lstlisting}

Here \texttt{With(...)} refines the canonical transducible function
\texttt{Answer $\ll$ Question} 
with the task-specific instructions and tools without changing the typed interface.
When explanation is set to True, the transducible function returns two arguments, the second of them returning explanations and evidence traces. 
\begin{lstlisting}[language=Python]
answer, explanation = await decide(q)
print(explanation)
{
  "explanation": "The option vibration was selected because ...",
  "relevant_source_attributes": [
    "prompt",
    "options"
  ],
  "confidence": 0.96
}
\end{lstlisting}

\subsection{Type Operators}

\paragraph{Type Composition with \texorpdfstring{@}{@}.}
In Agentics 2.0, the operator \texttt{@} performs state composition between two
Pydantic types or states. 
Given types \texttt{A} and \texttt{B}, the composed type \texttt{A @ B} is a new
type with two fields, 
\texttt{left: A} and \texttt{right: B}, used to keep an original object
and its derived state together.

\begin{lstlisting}[language=Python]
ComposedType = Answer @ Question
print(ComposedType.model_fields)
{'left': FieldInfo(annotation=Question),
 'right': FieldInfo(annotation=Answer)}
\end{lstlisting}

\paragraph{Type Merge with \texorpdfstring{\&}{\&}.}
The operator \texttt{\&} performs a merge of any two Pydantic Types or states. 
At the type level, \texttt{MergedType = Answer \& Question} produces a new type that combines the fields of both into a single field, 
preferring one side when assigning values from two instances. 
At the instance level, \texttt{\&} aggregates two compatible instances into one, with the one side taking precedence when both define a value.
\begin{lstlisting}[language=Python]
MergedType = Answer & Question
print(MergedType.model_fields)
{ 'prompt': FieldInfo(annotation=str, ...),
  'options': FieldInfo(annotation=list[str], ...),
  'choice': FieldInfo(annotation=str, ...),}
\end{lstlisting}

\subsection{Map-Reduce Semantics}

When invoked on a list of inputs of the required source types, 
transducible functions can either execute $map$ or $reduce$ operator 
over all the input states.
When executing the $map$ operator on the transducible function, all the input states are processed in parallel, and it returns a corresponding list of outputs, whose order reflects the original input order.
On the other hand, a $reduce$ operator accepts a list of states as input and returns a single state as output. 
If the input list is too large to fit into a single LLM prompt, Agentics 2.0 internally reduces it into parallel, asynchronous batches, which are then aggregated in stages to produce a single final state.
\begin{lstlisting}[language=Python]
questions: list[Question] = ...
answers: list[Answers] = await decide(contexts)
\end{lstlisting}

\subsection{Decorator \texttt{@transducible}}

Agentics 2.0 provides the ability to wrap an arbitrary asynchronous Python function into a transducible function using the \texttt{@transducible} decorator, provided they take a single Pydantic input state and return a single Pydantic output state. 
This feature provides a lightweight $map$-$reduce$ design pattern that facilitates scaling their execution. A transducible function can invoke multiple transductions, which makes it easier to develop highly composable and controllable workflows.
Additionally, Agentics 2.0 offers primitives that enable the creation of sophisticated behaviors where regular Python code and LLM-based transduction can be combined in a transducible function. This is made possible through the ability to return and manipulate Pydantic objects.
In the example below, the input state is manipulated (concatenated \texttt{send it to Agent}) 
and then sent to transduction (\texttt{Transduce}).
\begin{lstlisting}[language=Python]
from agentics import transducible
@transducible(provide_explanation=True)
async def write_an_email_to_alfio(state: GenericInput) -> Email:
    """Write an email about the provided content"""
    state.content=state.content + " send it to Agent"
    return Transduce(state)
\end{lstlisting}


%% file: discoverybench.tex
\section{DiscoveryBench}
\subsection{Task and Benchmark Summary}
\paragraph{Problem}
\texttt{DiscoveryBench}\footnote{\url{https://github.com/allenai/discoverybench}} formalizes data‑driven discovery as a benchmark task \citep{majumder2025discoverybench}.
Given one or more datasets in a comma-separated-value (CSV) file and metadata, the main goal is to derive a hypothesis 
$h$ as a function of context $c$, variables $v$, and relations $r$ that constitute $\psi(c, v, r)$.
This hypothesis function can express a declarative statement, "under context c, variables v have relationship r".
Solutions typically require multi‑step analysis pipelines that combine semantic reasoning for mapping required terms to columns and statistical analysis for model selection. 
The hypothesis in the problems has a tree structure, where a hypothesis consists of several sub-hypotheses over a path in the tree.

The benchmark comprises manually created DB-REAL and LLM-generated DB-SYNTH.
DB-REAL comprises 10 problem sets ranging from archaeology, requirement engineering, regression, etc \footnote{In our evaluation, we took 9 problem sets 
except for the meta-regression-raw, and excluded 2 problems in the world-bank-education-gdp domain. In total, we evaluated 186 problems.}.
In the benchmark results from \texttt{ASTABench} \citep{bragg2025astabench}, we see that the best reported final score is 33.7, reflecting significant challenges for solving this benchmark.

\paragraph{Evaluation}
The hypotheses are natural language sentences, and the evaluation protocol uses an LLM–as–judge approach to compute the Hypothesis Matching Score (HMS) or final score.
The score computation procedure first decomposes both the gold and predicted hypotheses into sub-hypotheses, then scores context alignment, variable alignment, and relationship accuracy, and finally combines the three scores and multiplies the result by 100 to get the final score between 0 and 100.

We use \texttt{Inspect AI} \citep{inspectai} to evaluate all agents, and we use the default \texttt{gpt-4o} model as the evaluation model for all experiments.

\subsection{Agentics 2.0 Implementation}
We summarize the common design patterns that leverage the Agentics 2.0 framework\footnote{Due to the space limit, we provide the full implementation in the supplementary material.}.
We first define types for transduction from the inintial input \textit{Question} 
to the final \textit{Answer} through \textit{IntermediateEvidence} types.
\begin{lstlisting}[language=Python, caption={Pydantic Types for Question, Answer, Evidence}]
class IntermediateEvidence(BaseModel):
    evidence_found: bool | None
    evidence: str | None
    partial_answer: str | None
class Answer(BaseModel):
    short_answer: str | None
    selected_evidence: list[IntermediateEvidence] | None
class Question(BaseModel):
    question: str | None
    question_type: str | None
    dataset: str | None
    metadata: dict | None
    dbs: list[AgenticDB] | None
\end{lstlisting}

The overall agentic data workflow is a two-stage pipeline. 
The first stage is to collect \textit{IntermediateEvidence} states from available sources
such as the structured data in CSV files and all the outputs from ReAct agent.

\begin{lstlisting}[language=Python, caption={Stage 1. Collects Evidence Direclty from Structured Data}]
async def collect_evidence_for_source(table_df, q: Question) -> list[IntermediateEvidence]:
    # Canonical type transduction from table
    data = AG.from_dataframe(table_df) 
    evidence = await (AG(
        atype=IntermediateEvidence,
        transduction_type="areduce", 
        instructions="You are extracting INTERMEDIATE EVIDENCE to answer the QUESTION later.") << data)
    return evidence.states
\end{lstlisting}
The second stage aggregates collected intermediate answers across multiple input sources into a single final answer.
\begin{lstlisting}[language=Python, caption={Stage 2. Reduce Evidence to Final Answer as a Hypothesis}]
async def synthesize_hypothesis(evidences: list[IntermediateEvidence], q: Question) -> Answer:
    ans = await (AG(
    atype=Answer, 
    transduction_type="areduce", 
    instructions="Given prior INTERMEDIATE EVIDENCE from multiple sources, produce one hypothesis.") << AG(atype=IntermediateEvidence, states=evidences))
    return ans[0] if len(ans)>0 else Answer()
\end{lstlisting}
Both stages can be called in a driver function per question as follows.
The function processes each question in parallel, and 
For each question, it collects multiple pieces of evidence from database sources, typically one to three csv files specified in the problem description. Then, the collected evidence is reduced to the final answer type with a hypothesis to be graded by LLMs.
\begin{lstlisting}[language=Python, caption={Map-Reduce Driver Function}]
async def answer_question_from_data(q: Question) -> Question:
    per_source = []
    for db in q.dbs:
        # map over sources
        per_source += await collect_evid_for_src(db.df, q)    
    # reduce across sources
    final = await synthesize_hypothesis(per_source, q)
    return q
\end{lstlisting}

Compared with the prompt-centric implementation of ReAct agents \citep{yao2022react},
the above agentic data workflow implementation in Agentics 2.0 leverages the advantage of the strongly typed transducible functions, and prompting is a part of a larger Python program. Note that the raw table imported from a data frame is also parsed into a Pydantic type directly via canonical type transduction without human annotation of the database.

\subsection{Evaluation Results}
We evaluate agents written in Agentics 2.0 and the baseline ReAct agent following the evaluation protocols of the ASTSA-Bench on DB‑REAL datasets. 
Agents written in Agentics 2.0 are wrapped as a solver of \texttt{agent-baselines} \citep{agentbaselines}
and we evaluate the baseline ReAct solver of the \texttt{agent-baselines} implementation.
In total, we evaluated four algorithm configurations: 
\texttt{baseline-react}, 
\texttt{agentics-agg} that reduces structured data into intermediate evidence,
\texttt{agentics-react} that maps the ReACT agent generations into intermediate evidence,
and 
\texttt{agentics-both} that collects intermediate evidence from both structured data and ReAct agent generations.

The score of the agents also depends on the generation model, and we 
evaluated all algorithm configurations on \texttt{gemini-3-flash-preview} and \texttt{gpt-4.1} models.
For each algorithm configuration, we run it at least 5 times to compute the mean and standard deviation of the scores.

\subsubsection{Aggregated Scores}
\begin{figure}[t]
     \centering
     \begin{subfigure}[b]{0.45\textwidth}
         \centering
         \includegraphics[width=\textwidth]{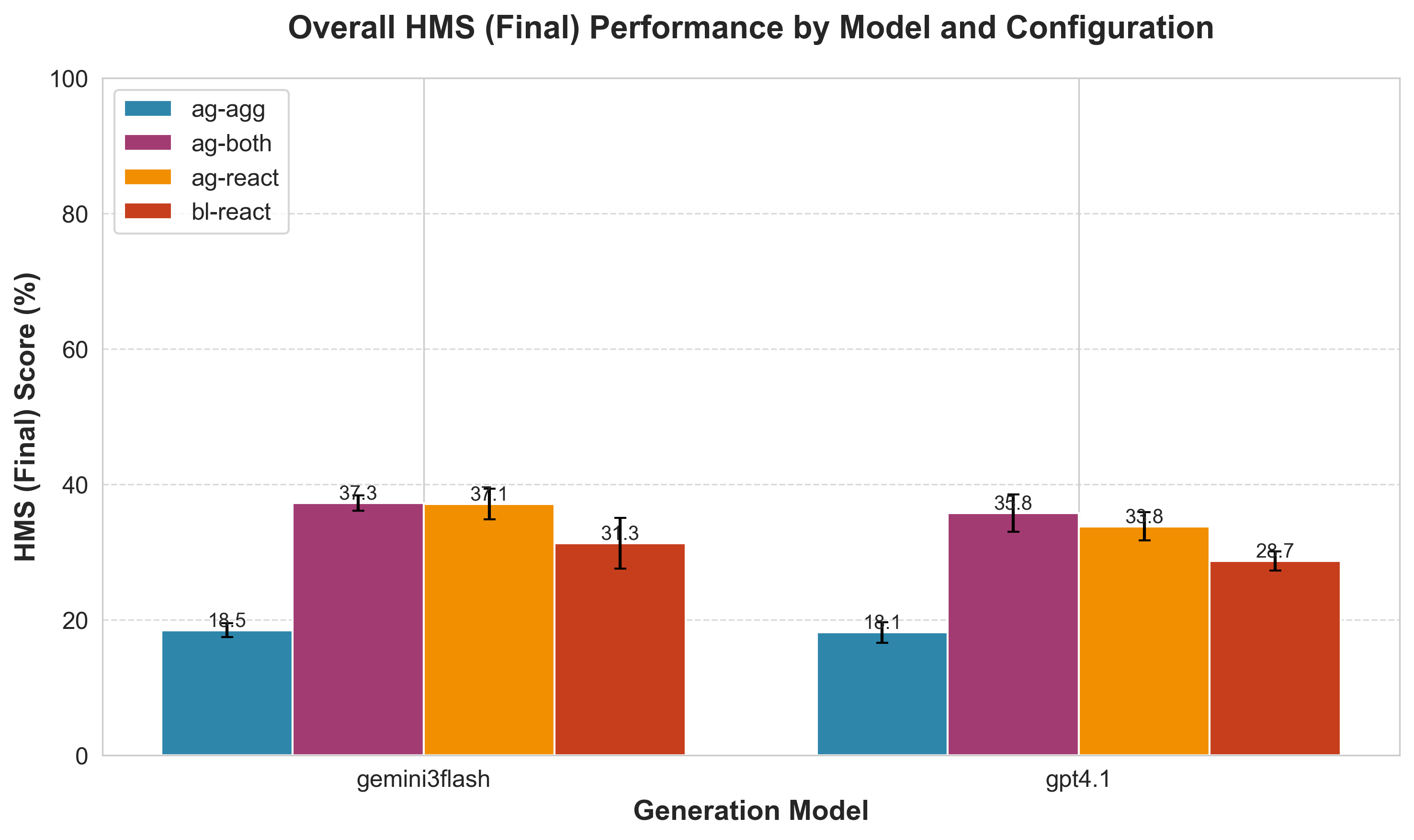}
         \caption{Hypothesis Matching Scores}
         \label{fig:overall-a}
     \end{subfigure}
     \hfill
     \begin{subfigure}[b]{0.45\textwidth}
         \centering
         \includegraphics[width=\textwidth]{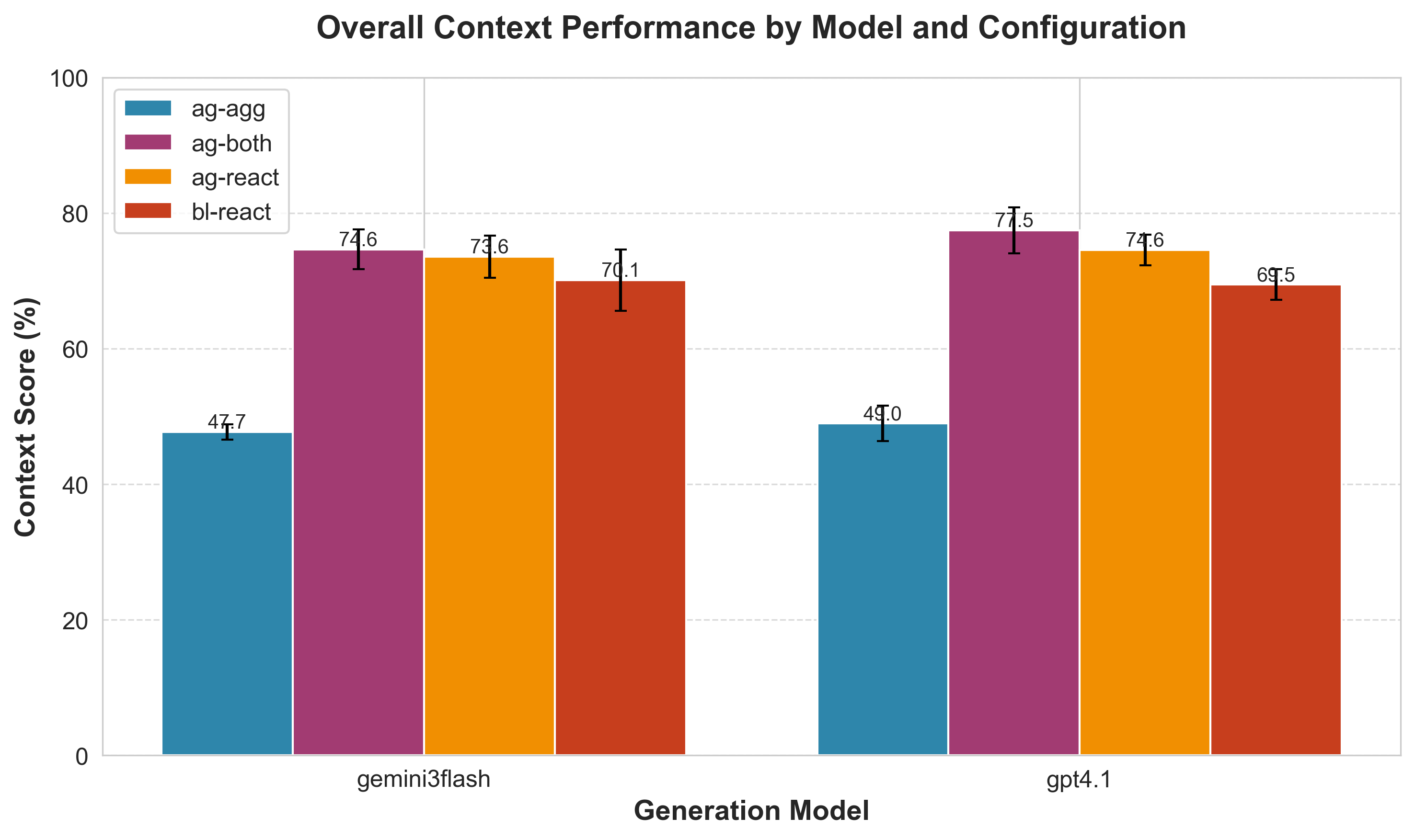}
         \caption{Context Scores}
         \label{fig:overall-b}
     \end{subfigure}
     \begin{subfigure}[b]{0.45\textwidth}
         \centering
         \includegraphics[width=\textwidth]{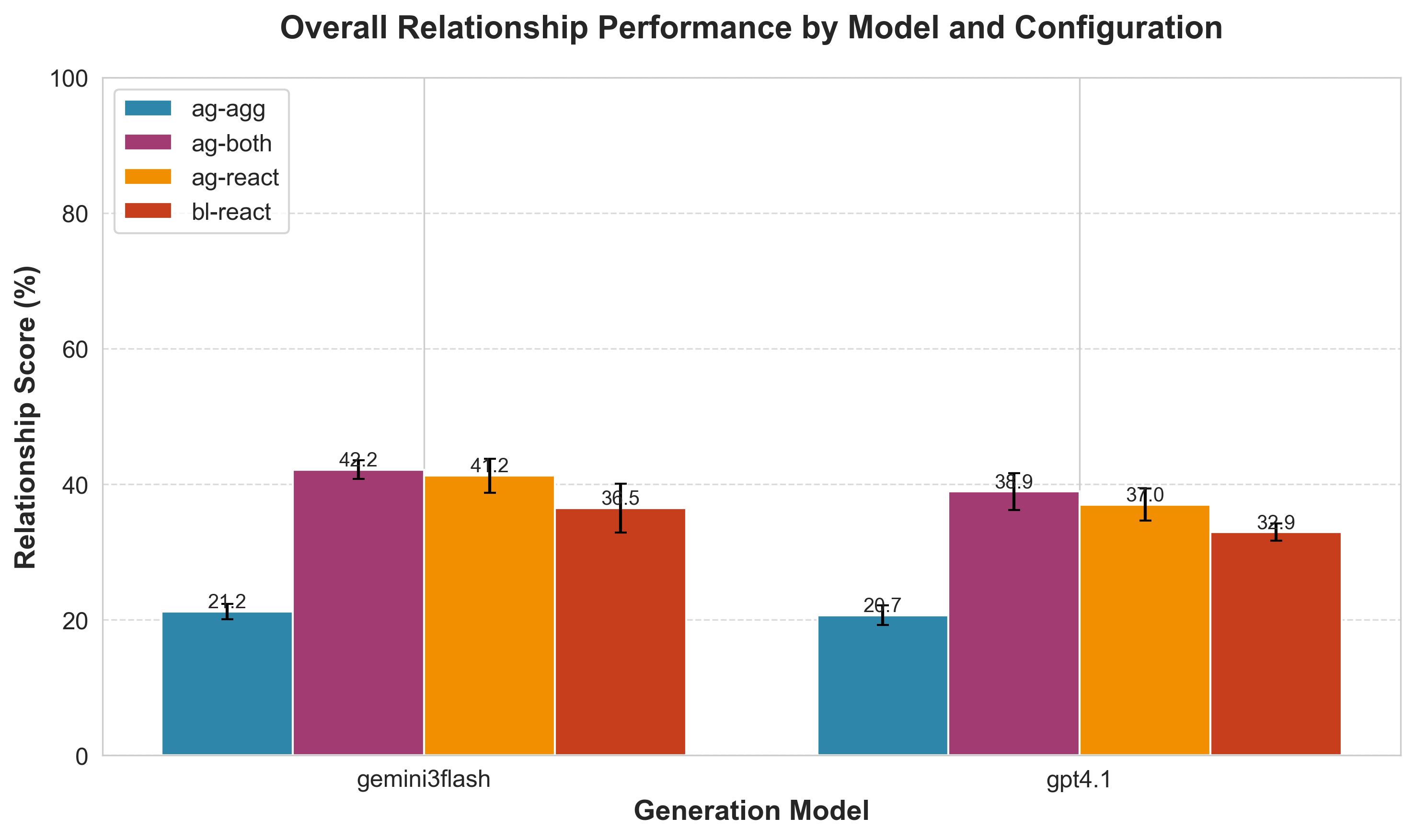}
         \caption{Relation Scores}
         \label{fig:overall-c}
     \end{subfigure}
     \hfill
     \begin{subfigure}[b]{0.45\textwidth}
         \centering
         \includegraphics[width=\textwidth]{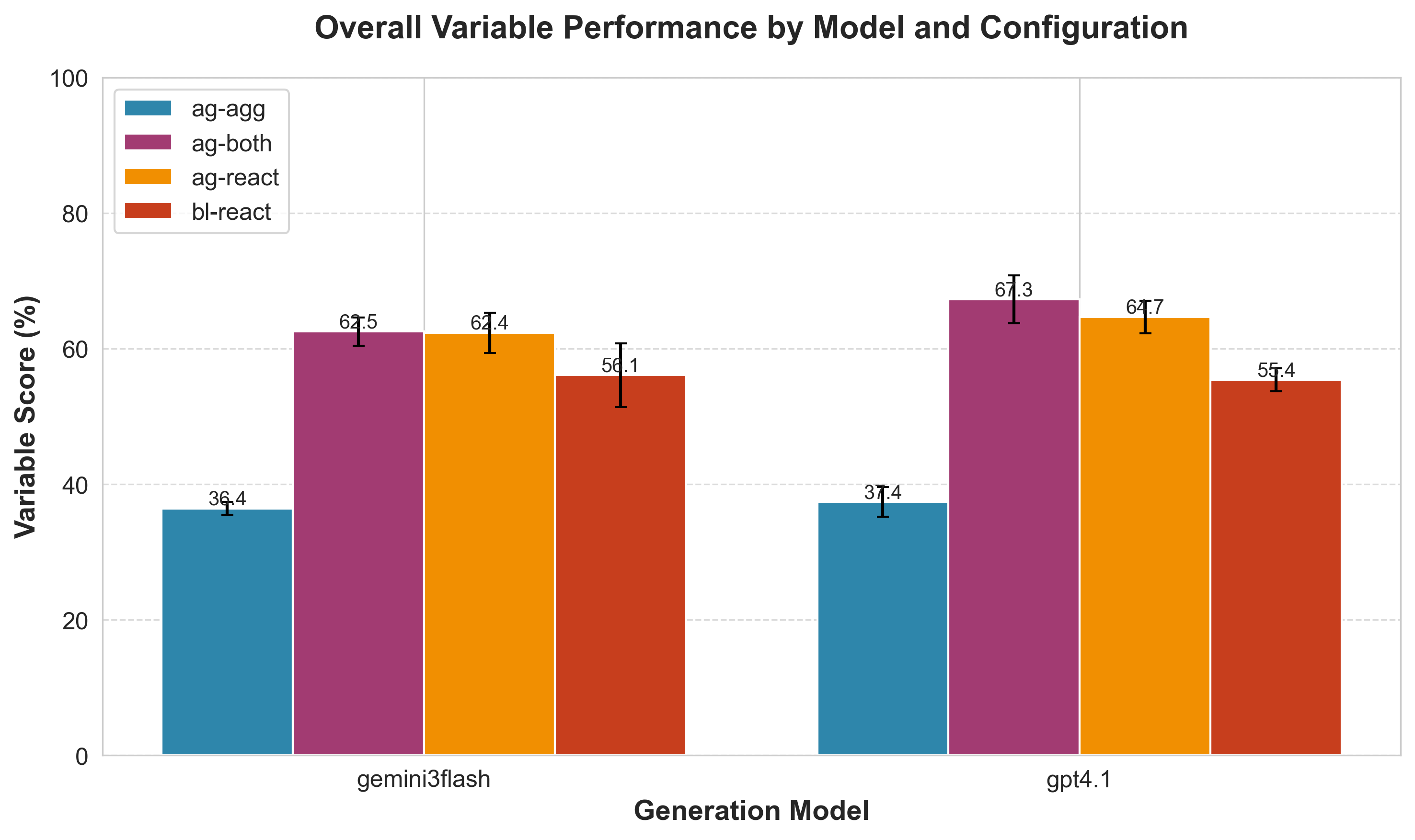}
         \caption{Variable Scores}
         \label{fig:overall-d}
     \end{subfigure}
     \caption{Overall aggregated scores.
     All figures show the aggregated scores over 10 problem domains,
     evaluating \texttt{agentics-agg} (ag-agg), 
     \texttt{agentics-both} (ag-both), 
     \texttt{agentics-react} (ag-react),  and
     \texttt{baseline-react} (bl-react) 
     using \texttt{gemini-3-flash-previuew} (gemini3flash) and \texttt{gpt-4.1} (gpt4.1) models.
     }
     \label{fig:overall-scores}
\end{figure}

Figure \ref{fig:overall-scores} shows summary of hypothesis matching scores,
context scores, relation scores, and variable scores 
of four algorithm configurations using two LLMs.
We see that  \texttt{agentics-both} and \texttt{agentics-react} are the two top performers among the four configurations. Note that \texttt{agentics-both} achieves the average final score \textbf{37.27} \footnote{This score is higher than the best score (33.7 by baseline ReAct) available in the leaderboard \url{https://allenai-asta-bench-leaderboard.hf.space/data-analysis}}.
The evaluation result of \texttt{baseline-react} is similar to the scores reported in the leaderboard \url{https://allenai-asta-bench-leaderboard.hf.space/data-analysis}.
When we look into the three scores from Figure \ref{fig:overall-b} to \ref{fig:overall-d},
All agents achieve relatively higher scores in extracting the context and variables, but struggle to extract the correct relations between variables.

\subsubsection{Per Dataset Scores}
\begin{figure}[t]
     \centering
     \begin{subfigure}[b]{0.45\textwidth}
         \centering
         \includegraphics[width=\textwidth]{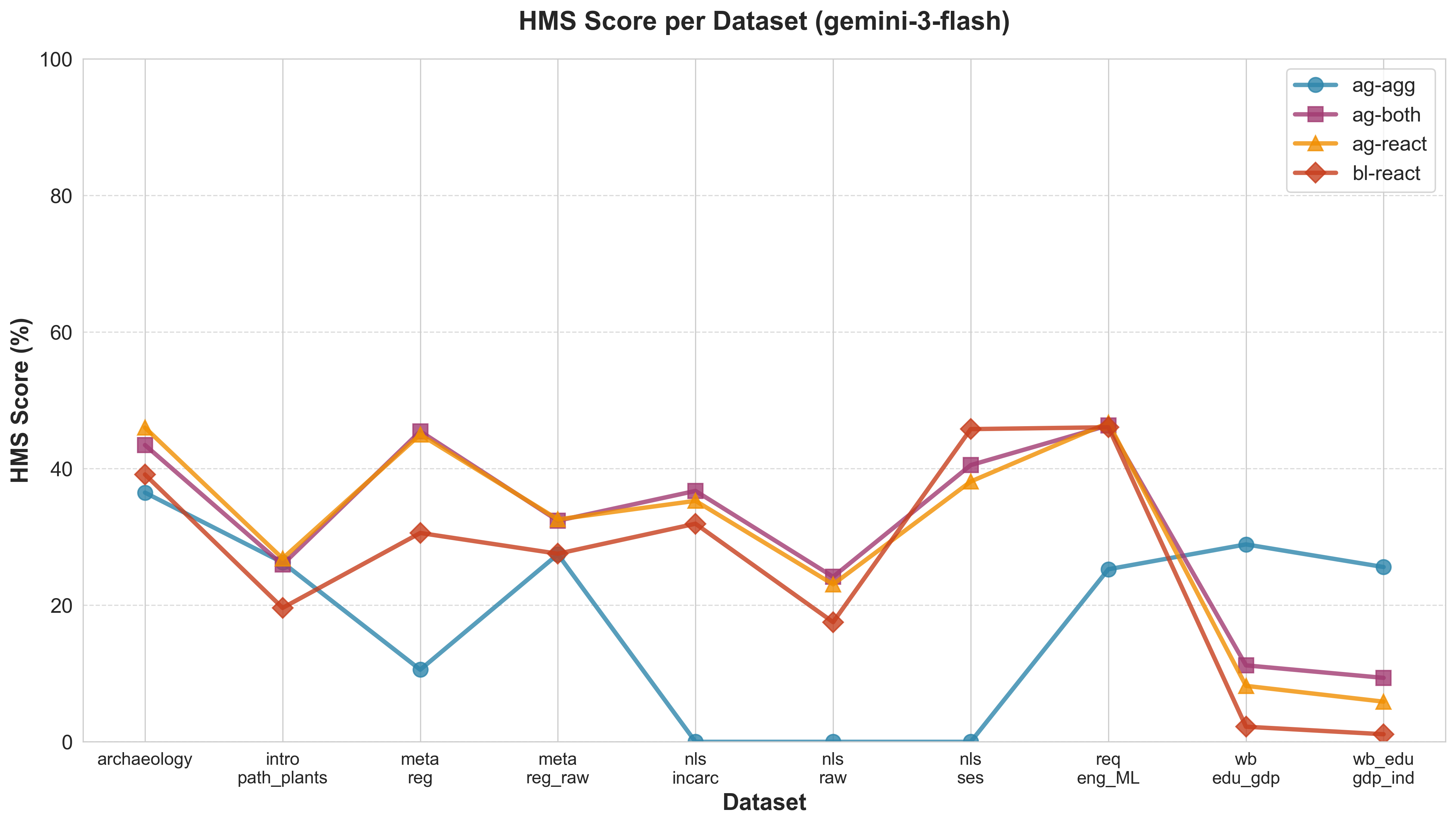}
         \caption{Gemini-3-Flash-Preview}
         \label{fig:data-a}
     \end{subfigure}
     \hfill
     \begin{subfigure}[b]{0.45\textwidth}
         \centering
         \includegraphics[width=\textwidth]{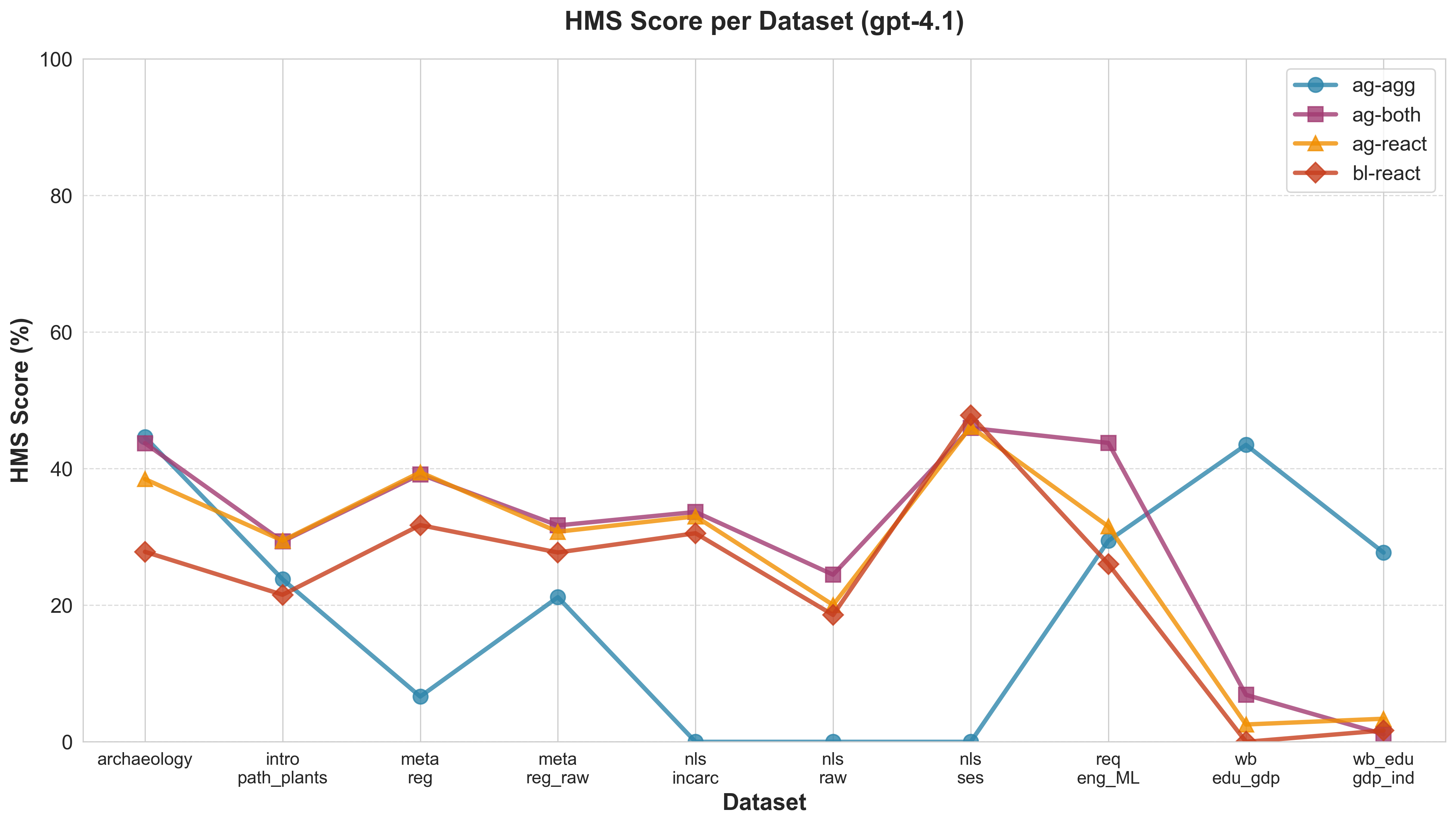}
         \caption{GPT-4.1}
         \label{fig:data-b}
     \end{subfigure}
     \caption{Aggregated scores per dataset.
     All figures show the aggregated hypothesis matching scores per each dataset,
     and show the results from four algorithm configurations
     \texttt{agentics-agg} (ag-agg), 
     \texttt{agentics-both} (ag-both), 
     \texttt{agentics-react} (ag-react),  and
     \texttt{baseline-react} (bl-react).
     }
     \label{fig:data-scores}
\end{figure}

Next, we break down the aggregated score for each of the ten datasets.
Figure \ref{fig:data-scores} shows the hypothesis matching scores 
from the four algorithm configurations.
We see that \texttt{agentics-both}, \texttt{agentics-react}, and \texttt{baseline-react}
show a similar pattern overall, only differing at the level of the scores, with \texttt{agentics-both} being the highest and \texttt{baseline-react} being the lowest.

\begin{table}[h!]
\centering
\footnotesize
\begin{tabular}{lrr}
\hline
\textbf{CSV File} & \textbf{Rows} & \textbf{Cols} \\
\hline
archaeology/capital.csv & 400 & 14 \\
archaeology/time\_series\_data.csv & 3600 & 37 \\
archaeology/pollen\_openness\_score\_...csv & 4084 & 3 \\
intro-path-plants/temporal\_trends\_contingency\_table.csv & 12 & 3 \\
intro-path-plants/invasion\_success\_pathways.csv & 81 & 19 \\
intro-path-plants/invaded\_niche\_pathways.csv & 632 & 25 \\
meta-reg/joined-data.csv & 88 & 59 \\
meta-reg-raw/success-data.csv & 928 & 50 \\
meta-reg-raw/study-data.csv & 143 & 15 \\
nls-incarc/processed.csv & 12014 & 6 \\
nls-raw/raw.csv & 12686 & 62 \\
nls-ses/ses-processed.csv & 8773 & 9 \\
req-eng-ML/req-eng-ml.csv & 276 & 214 \\
wb-edu-gdp/wb-edu-gdp.csv & 13 & 45 \\
wb-edu-gdp-ind/ndicator.csv (6 files) & 2 each & 2 each \\
\hline
\end{tabular}
\caption{Size of CSV Files. The number of rows and columns of the CSV files in each dataset.}
\label{tab:csv-size}
\end{table}
Note that \texttt{agentics-agg} performs well or is competitive to the other three configurations
on five data sets: \textit{archaeology}, \textit{introduction pathways non-native plants}, \textit{requirements engineering for ML enabled systems}, \textit{worldbank eduction gdp}, and \textit{worldbank education gdp indicators}.
This implies that LLMs can extract context, variables, and relations solely from table inputs with additional metadata, without fitting any machine learning models to those problem sets.

Table \ref{tab:csv-size} shows the size of CSV files in each dataset.
We see that \texttt{agentics-agg} failed to solve a single problem in 
\textit{nls-incarceration},  \textit{nls-raw}, and  \textit{nls-ses}, 
which has large number of rows.
On the other hand, 
 \textit{worldbank eduction gdp}, and \textit{worldbank education gdp indicators} 
 have only two rows and two columns over 6 files, and
 \textit{requirements engineering for ML enabled systems} has 276 rows and 214 columns.
 This indicates that \texttt{agentics-agg} can aggregate the input table 
 to extract meaningful intermediate evidence when the size of the table is manageable.

\subsubsection{Per Question Type Scores}
\begin{figure}[h]
     \centering
     \begin{subfigure}[b]{0.3\textwidth}
         \centering
         \includegraphics[width=\textwidth]{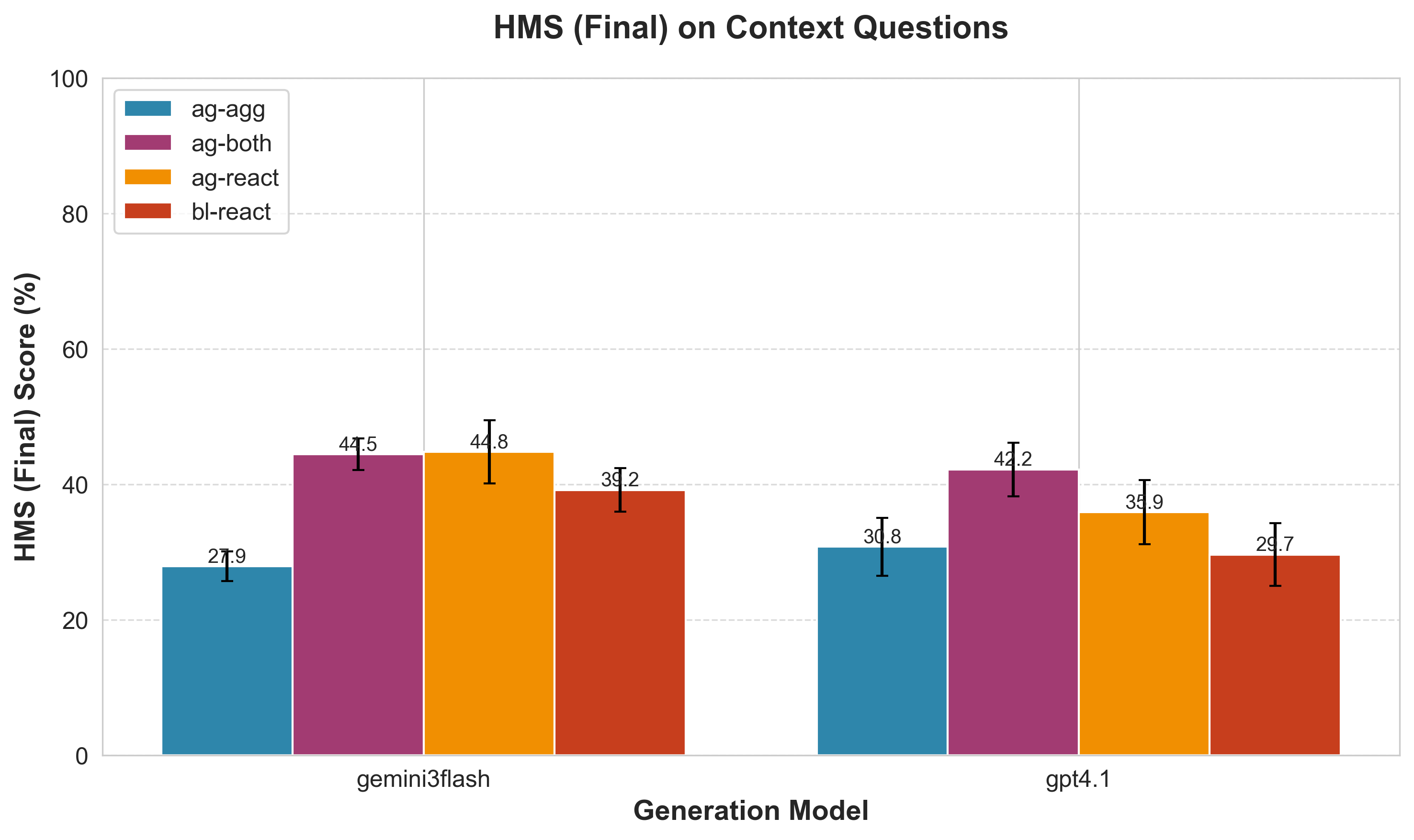}
         \caption{Context questions}
         \label{fig:question-a}
     \end{subfigure}
     \begin{subfigure}[b]{0.3\textwidth}
         \centering
         \includegraphics[width=\textwidth]{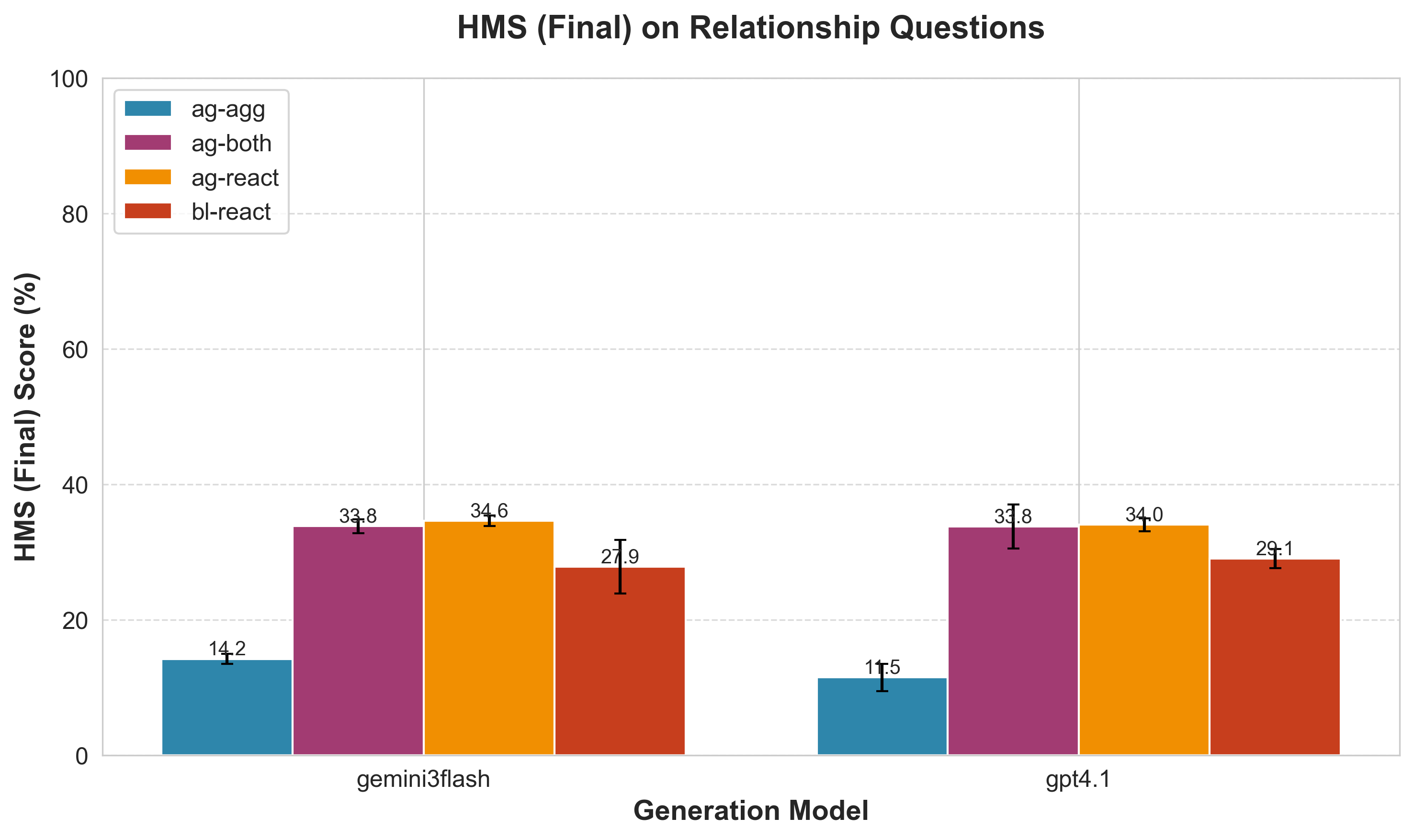}
         \caption{Relationship questions}
         \label{fig:question-b}
     \end{subfigure}
     \begin{subfigure}[b]{0.3\textwidth}
         \centering
         \includegraphics[width=\textwidth]{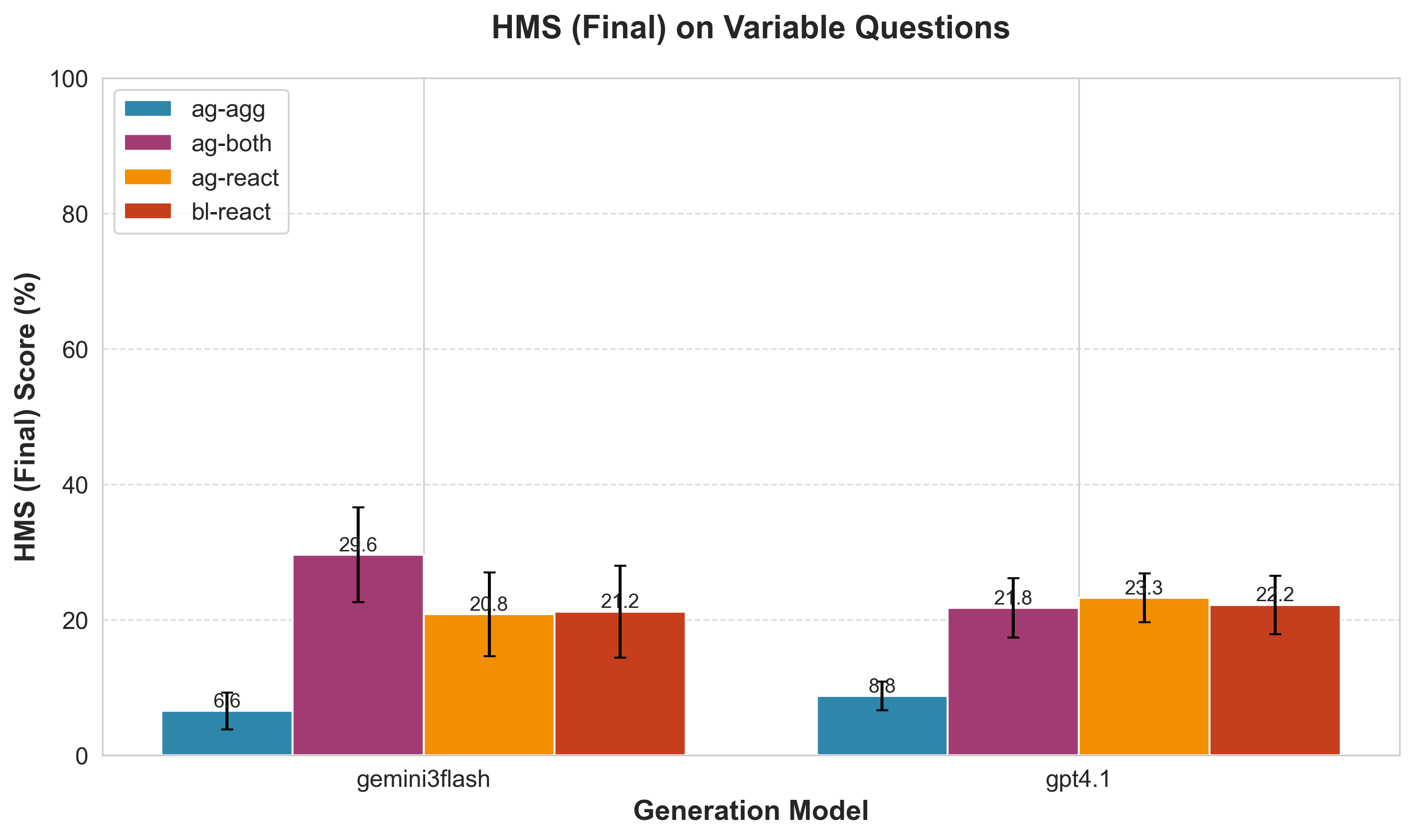}
         \caption{Variable questions}
         \label{fig:question-c}
     \end{subfigure}
     \caption{Aggregated scores per question types.
     All figures show the aggregated scores for three question types: context questions, relationship questions, and variable questions.}
     \label{fig:question-scores}
\end{figure}

Figure \ref{fig:question-scores}
shows the hypothesis-matching scores across three question types.
We observe that all four algorithm configurations perform best on context questions and worst on variable questions.
This indicates that the task of generating the question by hiding the variables is the most difficult of the three. 
Especially \texttt{agentics-agg} shows signicant drop from 
\textbf{27.9} in context questions to \textbf{6.6} in variable questions when it uses \texttt{gemini-3-flash-preview} model.
On the other hand, \texttt{agentics-agg}
remains relatively robust to the change of the question types
compared with the other three, dropping from \textbf{44.5} to \textbf{29.6}.

\subsection{Take Away}
Since our paper presents the Agentics 2.0 framework, we didn't incorporate benchmark-specific techniques and focused on developing agents with transducible functions over structured data.
First, we see that agents implemented in Agentics 2.0 consistently outperform the best-known baseline in the leaderboard.
Second, when the dataset is small, aggregating tables performs well without generating code to fit models.
Third, all agents perform relatively better on context questions or achieve higher context scores, but they struggle to solve variable questions and show lower scores in extracting the correct relations between variables.

%% file: archer.tex
\section{Archer NL to SQL}
\begin{figure}[t]
  \centering
  \includegraphics[width=0.4\linewidth]
  {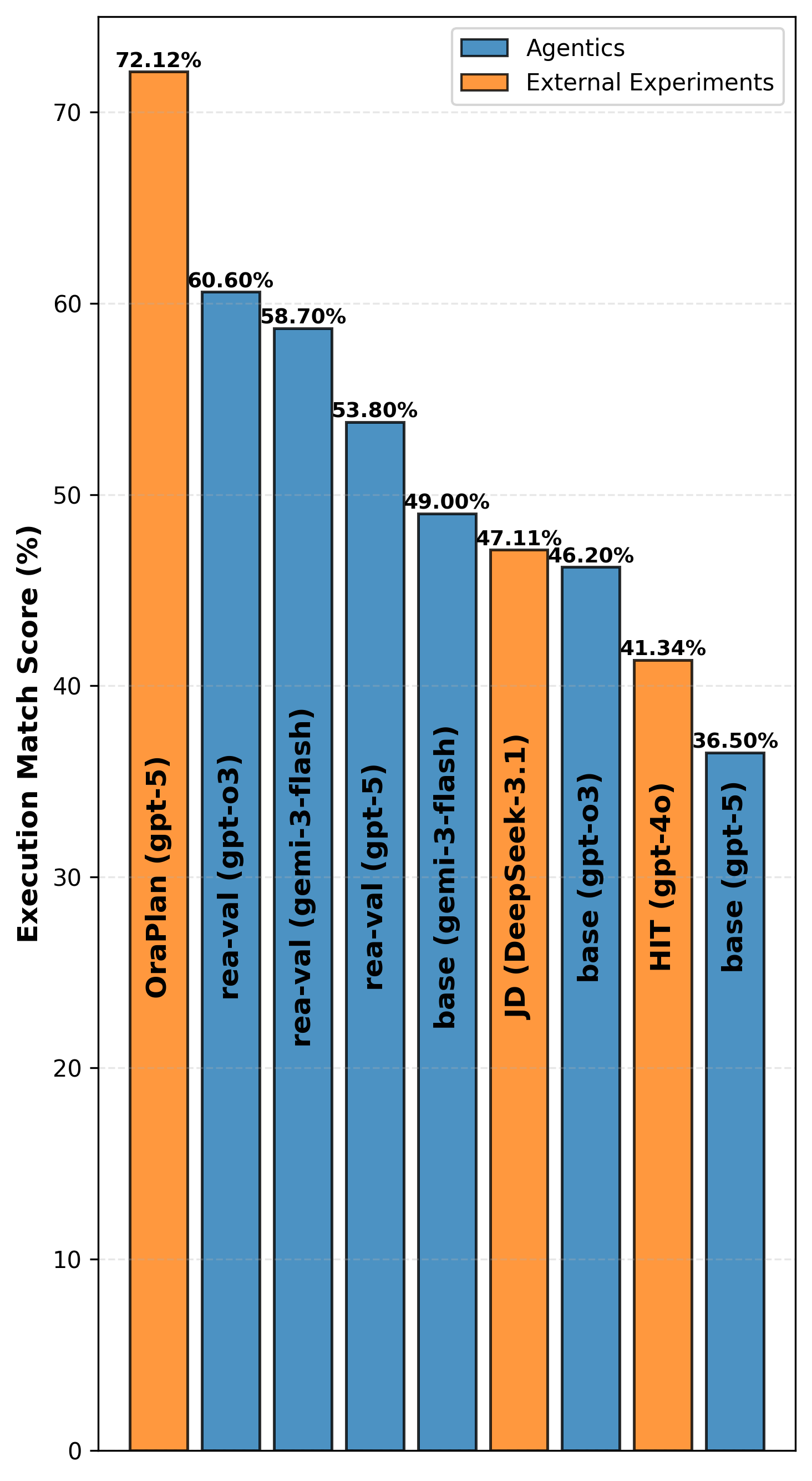}
  \caption{Archer English Devset Execution Match. 
  The figure shows the aggregated execution match score on the English devest, and shows the results from  Agentics 2.0 implementation in blue bars and the Archer leaderboard results in orange bars.
  }
  \label{fig:archer-execution-match}
\end{figure}

\subsection{Task and Benchmark Summary}
\paragraph{Problem}
Archer is a bilingual natural language (NL) to structured query language (SQL) benchmark designed to test 
reasoning capabilities along with the NL to SQL semantic parsing task \cite{zheng2024archer}. 
Semantic parsing from NL to SQL is extremely challenging compared with mathematical or common-sense reasoning tasks. Indeed, the leaderboard of Archer \footnote{\url{https://sig4kg.github.io/archer-bench/}} still shows that the best test score is merely 54.96 from OraPlan-SQL \cite{liu2025oraplan}.
Compared with other NL to SQL benchmarks \cite{li2023can}, 
Archer requires arithmetic, common sense, and hypothetical reasoning to derive the database entities or expressions to capture the desired intent in the question. 

\paragraph{Evaluation}
Following the most popular evaluation protocol for NL to SQL, we compute the execution match score by comparing the result sets returned by the gold and the predicted SQL queries.
Namely, the score is 1 if both result sets match and 0 otherwise. In addition, we report a relaxed metric that computes the F1 score by comparing both result sets, assigning a partial score if the predicted result set contains partially correct results.

\subsection{Agentics Implementation}
To address the challenges of reproducing leaderboard baseline scores, we implemented two agents in Agentics 2.0. The basic agent generates an SQL query from the given question and verifies its correctness against the database engine, providing feedback if it contains any syntax errors.

The reasoning-validation agent follows a more reasoning-intensive workflow. It begins by selecting natural language and SQL pairs from the training examples that share the same reasoning type. Next, it analyzes the extracted few-shot dataset to understand the task and formulate a strategy to address the reasoning challenges in SQL generation. After generating a syntactically verified SQL query, it performs an additional reasoning step to validate its semantics by checking the previously developed strategies and thoughts generated in the previous iterations.

\subsection{Evaluation Result}
Figure \ref{fig:archer-execution-match} shows the execution match scores from Agentics 2.0 implementations and the leaderboard results. The rea-val denotes the agent that performs reasoning and semantic validation, and base denotes the agent that performs basic generation and verification. 
We see that the NL to SQL agents implemented in Agentics 2.0 outperform all leaderboard submissions except for the OraPlan-SQL \cite{liu2025oraplan}, which 
involves a sophisticated benchmark-specific strategy for domain analysis and planning.

Table \ref{tab:archer-reasoning-type} shows the detailed scores aggregated per reasoning type for the reasoning-validation agent. We can see that GPT-O3 outperformed the Gemini-3 Flash model in the reasoning category that requires both arithmetic and commonsense reasoning. 

\begin{table}[h]
\centering
\begin{tabular}{l l r r r}
\toprule
\textbf{Reasoning} & \textbf{Model} & \textbf{EX} & \textbf{F1} & \textbf{Count} \\
\midrule
A     & Gemini-3-flash & 0.875 & 0.875 & 24 \\
A     & GPT-o3         & 0.833 & 0.872 & 24 \\
\midrule
A C   & Gemini-3-flash & 0.393 & 0.474 & 28 \\
A C   & GPT-o3         & \textbf{0.607} & \textbf{0.674} & 28 \\
\midrule
A H   & Gemini-3-flash & \textbf{0.682} & \textbf{0.808} & 22 \\
A H   & GPT-o3         & 0.591 & 0.682 & 22 \\
\midrule
A C H & Gemini-3-flash & 0.467 & 0.467 & 30 \\
A C H & GPT-o3         & 0.433 & 0.515 & 30 \\
\bottomrule
\end{tabular}
\caption{Archer NL to SQL by Reasoning Type.
The score shows the result from the reasoning-validation agent using two best-performing models, Gemini-3-flash and GPT-O3. A, C, and H refer to arithematic, commonsense, and hypothetical reasoning, respectively.
}
\label{tab:archer-reasoning-type}
\end{table}

\subsection{Take Away}
Agents developed within Agentics 2.0 perform competitively compared to submissions on the Archer leaderboard. 
The NL to SQL agent involves multiple stages of problem analysis, SQL query execution, and several iterations of generation, verification, and refinement. 
It may also use the external tools, such as a schema linker to extract a relevant view of the database to shorten the context length of the prompt, or a database query engine to sample the database to gain a better understanding of its content.
Similar to DiscoveryBench, we see a promising direction for the Agentics 2.0 framework to extend its implementation by incorporating problem-specific techniques to further improve performance.

%% file: conclusion.tex
\section{Discussion}
\paragraph{Advantages.}
Our experiments show that developing agentic data workflows with transducible functions demonstrates several advantages.
The input and output types can be validated early to catch errors caused by failures in LLM inference calls or by decoding at the boundary between steps.
The evidence traces provide justification for the predicted slots,
the algebra of transducible functions, and the Map-Reduce model enable reusable design patterns.
The asynchronous workload orchestraion leverage the backend parallelism.
\paragraph{Limitation and Future Works.}
Our work has the following limitations.
The evidence formalization is still at the high-level and it requires richer logical systems to improve the fidelity of reasoning. Second, our current implementation assumes a single LLM backend, and integrating heterogeneous models and cost-aware scheduling is left for future work.
Third, while the present experiments span challenging benchmark domains such as hypothesis discovery and NL to SQL semantic parsing, broader evaluations and techniques tailored to each domain are desired.

\section{Conclusion}
We introduced Agentics 2.0, a Python-native framework for building type-safe, explainable, and
scalable agentic data workflows grounded in the logical transduction algebra.
By modeling LLM transduction inference as a sequence of transducible functions and leveraging the asynchronous Map-Reduce programming model, the proposed framework reconciles the flexibility of agentic systems with the rigor of typed semantics and logical evidence.

Through the design patterns that leverage asynchronous transducible functions, we have shown that a generic agent based on transducible functions can be implemented to achieve competitive performance against benchmark-specific state-of-the-art agents.


%% file: supp.tex
\section{AI Usage Disclosure}
LLMs were used for language polishing and LaTex formatting, fixing typos and punctuation. 
The code assistant LLMs are used to generate LaTex code for tables and 
Python scripts for generating figures using \texttt{matplotlib} Python package.
All technical content in the paper was written by the authors.